\documentclass{article}

\PassOptionsToPackage{numbers, sort&compress}{natbib}

\usepackage[preprint]{neurips_data_2024}

\usepackage[utf8]{inputenc} 
\usepackage[T1]{fontenc}    
\usepackage{hyperref}       
\usepackage{url}            
\usepackage{booktabs}       
\usepackage{amsfonts}       
\usepackage{nicefrac}       
\usepackage{microtype}      
\usepackage{xcolor}         
\usepackage{times}    
\usepackage{xspace}
\usepackage{graphicx}
\usepackage{amsmath}
\usepackage{amssymb}
\usepackage{booktabs}
\usepackage{spverbatim}
\usepackage{caption}
\usepackage{subcaption}
\usepackage[inline,shortlabels]{enumitem}
\usepackage{multirow}
\usepackage{hyperref}

\hypersetup{colorlinks=true,
    linkcolor=blue}      

\newcommand{\dataset}{OVR\xspace}
\newcommand{\modelname}{OVRCounter\xspace}

\title{OVR: A Dataset for Open Vocabulary Temporal Repetition Counting in Videos}

\author{
Debidatta Dwibedi \And Yusuf Aytar \And Jonathan Tompson \And Andrew Zisserman \and \{debidatta, yusufaytar, tompson, zisserman\}@google.com\\
Google Deepmind\\
}
\begin{document}

\maketitle

\begin{abstract}
We introduce a dataset of annotations of temporal repetitions in videos. 
The dataset, {\em \dataset} (pronounced as \textit{over}),  contains annotations for over 72K videos, with each annotation specifying 
the number of repetitions, the start and end time of the repetitions, and also a free-form description of what
is repeating.  The annotations are provided for videos sourced from Kinetics and Ego4D, and consequently
cover both Exo and Ego viewing conditions, with a huge variety of actions and activities. 
Moreover, \dataset  is almost an order of magnitude
larger than previous datasets for video repetition.
We also propose a baseline transformer-based counting model, {\em \modelname}, 
that can
localise and count repetitions in videos that are up to 320 frames long.
The model is trained and evaluated on the \dataset dataset, and its performance assessed with and without
using text to specify the target class to count. The performance is also compared to a prior repetition counting model. The dataset is available for download at:  \href{https://sites.google.com/view/openvocabreps/}{https://sites.google.com/view/openvocabreps/}
\end{abstract}    
\section{Introduction}
\label{sec:intro}

Whether it is cutting an onion or a morning workout, activities with
temporal repetitions are ubiquitous in our daily lives. Even solutions
to our most basic needs such as navigation, communication and
nourishment are handled by repetitive behaviours such as walking,
talking, and eating. Moreover, many dexterous human activities involve
deliberately repeating certain procedures to reach a goal one step at
a time (e.g.\ carving wood, painting, playing
guitar, stirring tea, cleaning surfaces etc.).

In this paper we introduce a new, large, open-vocabulary, video dataset for  temporal repetition counting.
We developed this Open-Vocabulary Repetitions  (\emph{\dataset}) dataset to enable algorithms to be trained and their performance evaluated for more
general counting capabilities than are possible at the moment. Our aim is a high quality {\em diverse} dataset. We achieve this by not restricting the vocabulary to a small set of classes. We ensure diversity by annotating existing datasets (Kinetics~\cite{kay2017kinetics} and Ego4D~\cite{grauman2022ego4d}) that have a huge variety of 
actions and activities. Since our aim is diversity and scale in the number of video clips, rather than scale in the duration of the repetition, the annotations cover up to 10s of video providing free-form textual description and the temporal extent of the repetition.

\begin{figure}[t]
\begin{center}
   \includegraphics[width=0.9\linewidth]{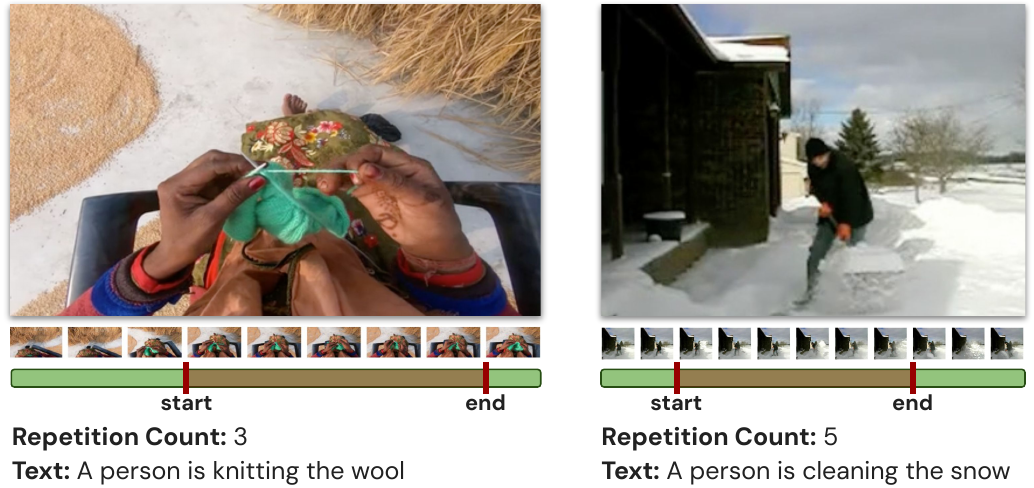}
\end{center}
\vspace{-2mm}
   \caption{\textbf{Samples from \dataset dataset}. Our data is annotated with {\bf open-vocabulary text descriptions} enabling text-conditioned repetition counting. \dataset dataset goes beyond sports and workout videos and introduces significant diversity and richness in the wild both from {\bf first-person} (left) and {\bf third-person} (right) views, at scale.}
\label{fig:dataset_teaser}
\end{figure}

\noindent\textbf{Why is a new dataset required?}
As has been demonstrated time after
time in many tasks (e.g.\ image understanding and video action
recognition), developing a new capability is only possible with proper datasets and
benchmarks guiding the way. For open vocabulary repetition counting, the dataset must be {\em large-scale} and {\em diverse}, as these are the dominant factors for determining the performance of a model on a given task.
However,
the existing temporal repetition counting datasets are quite limited
in size and diversity; See Table~\ref{tab:dataset_stats_final}. QUVA~\cite{runia2018real} and
UCFRep~\cite{zhang2020context}, both of which are noticeably diverse
repetition counting datasets, but have only $100$ and $526$ videos,
respectively. The RepCount~\cite{hu2022transrac} dataset increases the
amount of annotated data to roughly $1.5$K,  at the expense of
decreasing the diversity in the data distribution by limiting
repetitive activities to sports and workout domains. By sub-sampling
video clips from Kinetics~\cite{kay2017kinetics} action recognition
benchmark, the Countix~\cite{dwibedi2020counting} dataset annotates
$8.7$K videos with diverse human activities. This is almost an order
of magnitude larger than others yet still needed to be supported by a
synthetically generated dataset in order to train their model,
RepNet\cite{dwibedi2020counting}, to reach its best performance.

Our
dataset, \dataset, brings another order of magnitude increase in
sample size over Countix~\cite{dwibedi2020counting} while increasing diversity even further. This
will not only boost the performance of existing repetition counting
models, such as RepNet\cite{dwibedi2020counting} and
TransRAC\cite{hu2022transrac}, but it will also 
enable much larger models to be trained, potentially bringing 
superior performance. In addition to increasing the diversity of
third-person-view videos (by annotating all classes of Kinetics, beyond the set annotated in Countix), in the \dataset dataset we
annotate a significant number of first-person-view repetitive activities
by performing a large-scale crowdsourcing effort utilizing the
Ego4D~\cite{grauman2022ego4d} dataset where all the challenges coming
from ego-centric perception are naturally included. Annotated Ego4D
repetitions also bring diversity in geographical locations and
cultural settings as these are inherent qualities of the dataset.

\begin{table}
\setlength{\tabcolsep}{0.3em}
\centering

\renewcommand{\arraystretch}{1.2}{
    \begin{tabular}{l|r|c|c|r c c|r c r}

     & {\textbf{Number}} & \textbf{Free-form } & \textbf{Camera} & \multicolumn{3}{c|}{\textbf{Duration (s)}} & \multicolumn{3}{c}{\textbf{Count}} \\
    \cline{5-10}
     & \textbf{ of videos} & \textbf{Text} & \textbf{Viewpoint} & \textbf{Avg} & \textbf{Min} & \textbf{Max} & \textbf{Avg} & \textbf{Min} & \textbf{Max} \\
    \midrule
    \textbf{QUVA}~\cite{runia2018real} & 100~~~  & & Exo & 17.6 & 2.5 & 64.2 & 12.5  & 4  & 63\\
    \textbf{UCFRep}~\cite{zhang2020context} & 526~~~  &  & Exo & 8.2  & 2.1 & 33.8 & 6.7 & 3  & 54\\
    \textbf{RepCount}~\cite{hu2022transrac} & 1451~~~  & & Exo&29.4  & 4.0  & 88.0 & 15.9  & 1  & 141\\
    \textbf{Countix}~\cite{dwibedi2020counting} & 8757~~~  & &  Exo&6.1 & 0.2  & 10.0 & 6.8  & 2  & 73\\
    \textbf{\dataset (ours)} & {\bf 72552}~~ & \textbf{\checkmark} & \textbf{Exo + Ego}&2.7 & 0.03 & 10.0 & 4.7 & 2 & 121\\
    \end{tabular}}
\vspace{2mm}   
\caption{
{\bf Statistics of video repetition counting datasets.} Note that our dataset, \dataset, is almost an order of magnitude larger than the largest dataset available. The diversity is also expanded by exploring much wider range of contextual settings including different geographical locations in the world.  
\label{tab:dataset_stats_final}
}   
\end{table}

\noindent\textbf{Why open-vocabulary?}
Open-vocabulary semantics enables studying particular type of
repetitions using unrestricted language. For instance, localizing and
counting multiple different type of repetitions across time and space
would be useful for creating detailed summaries of multiple exercise
workouts from long videos. Being able to retrieve semantically similar
repetitions would enable robots to learn from relevant
first-person-recorded human activities instantly in the given
contextual setting. Open-vocabulary semantics not only gives meaning to
repetitions themselves but also allows us to study the way the
repetitions are done. For instance a sudden decrease in period length
in a basketball setting might correspond to an event change,
or an increase in period length in a bench press workout
might mean the athlete is struggling to put the weights up again. Understanding the
repetitive actions at a more detailed level will enable machines to
better understand the subtle temporal nuances in human actions.

Another significant benefit is that it allows {\em text-conditioned counting models} to be developed. 
We train a powerful baseline model, \modelname, that can be conditioned on the text description of a repeating action. Unlike class-agnostic repetition counting models like RepNet~\cite{dwibedi2020counting} and TransRAC~\cite{hu2022transrac} which count any repeating action, we can now control the counter to focus on only the specified action of interest. This unlocks the new capability of counting multiple classes of actions in the same video either spatially or temporally by using different text prompts.

Our main contribution in this paper is the OVR dataset and benchmark that is large scale, diverse and annotated with open-vocabulary text descriptions. We also train a powerful baseline model, \modelname, for text-conditioned counting which can also operate in a class-agnostic setting.     
\section{Related Work}
\label{sec:related_work}

\noindent\textbf{Temporal Repetition Counting.}
Earlier literature in this space \cite{tsai1994cyclic, cutler2000robust, vlachos2005periodicity, briassouli2007extraction, azy2008segmentation, pogalin2008visual} worked on understanding repetitions from the signal period using several types of crafted features, and were mainly studied in limited diversity settings. More recent work \cite{levy2015live, zhang2020context,runia2018real,dwibedi2020counting} increased diversity and studied temporal repetition counting more in the wild settings. While \cite{runia2018real} applied optical flow based methods, many others \cite{levy2015live,zhang2020context,dwibedi2020counting} employed deep learning methods. \cite{levy2015live} learned from synthetic sequences using CNNs. \cite{zhang2020context} introduced a network that utilizes temporal context at multiple scales and trained on a small annotated dataset (UCFRep). RepNet \cite{dwibedi2020counting}, a deep network architecture that utilizes temporal self-similarity matrices as bottlenecks, learned from synthetic repetitions generated from Kinetics videos, and finetuned the network on the limited data (Countix) that is annotated for temporal repetitions. Most recently \cite{zhang2021repetitive} utilizes sound in addition to frames, \cite{li2023full} employs temporal convolution networks over offline extracted features, and TransRAC \cite{hu2022transrac} introduces a deep learning model which processes data at multiple scales thereby achieving better performance. Improvements in performances are also achieved through specialized models that use human body poses (e.g.\ PoseRAC\cite{yao2023poserac}) but they may not generalize to in-the-wild settings where humans don't exist or are partially visible.

\noindent\textbf{Temporal Repetition Counting Datasets.}
Earlier datasets \cite{levy2015live,runia2018real} in this domain are quite small (100 videos) and only used for evaluation purposes. Later on  UCFRep~\cite{zhang2020context} and Countix~\cite{dwibedi2020counting}, with 526 and 8.7K videos respectively, are used for training deep models but their scale is still limited such that they require carefully designed network architectures and synthetic data for training. RepCount \cite{hu2022transrac}, with roughly 1.5K videos, increased the length of videos but limited the diversity to sports and workout videos. The summary statistics of these datasets can be viewed in Table~\ref{tab:dataset_stats_final}. Compared to existing datasets, our dataset offers a lot more diversity and scale, together with free-form text annotations.

\noindent\textbf{Semantics in Temporal Repetition Counting Datasets.} 
Temporal repetition counting in the wild is mostly explored in a
class-agnostic setting
\cite{runia2018real,dwibedi2020counting,zhang2020context} where the
target is counting repetitive behaviours independent of its
semantics. 
In fact, a few recent datasets
\cite{dwibedi2020counting,zhang2020context,hu2022transrac} already
have category-level labels as they are generated from  subsets of
existing action recognition benchmarks such as
Kinetics\cite{kay2017kinetics} and
UCF101\cite{soomro2012ucf101}. Although category labels create
opportunities to study class-specific repetition counting methods, they are not rich enough to enable open-vocabulary
understanding of repetitive activities in the wild. With \dataset dataset we open that path where less engineered deep models can be trained with open-vocabulary semantics for understanding the semantic context of repetitions and their attributes.  

\noindent\textbf{Counting Repetitions in Images.} The related task of counting `repeating' objects in images, e.g.\ counting cells or people, has seen a more rapid growth in challenge and interest, than that of temporal repetition counting. Early works had datasets for only a single class of
objects, and models that were only able to count a single class such as cells~\cite{Lempitsky10b}
or cars~\cite{mundhenkLargeContextualDataset2016a} . This
progressed to models that were `class-agnostic', able to count any
class, where the class was specified by providing `image exemplars'
(image patches from the target image covering the object of
interest)~\cite{dave,Gong2022ClassAgnosticOC, Liu22, Lu18,
10.1007/978-3-031-20044-1_20, m_Ranjan-etal-CVPR21,
Shi2022RepresentCA, yangClassagnosticFewshotObject2021, You_2023_WACV,
low_shot, Lin_2022_BMVC}. Research in this area was encouraged by the release of datasets, such as FSC-147~\cite{m_Ranjan-etal-CVPR21},
that covered a large variety of classes. Recent work has moved onto class-agnostic models where the
target class is specified by a text description~\cite{AminiNaieni23,Jiang2023CLIPCountTT, Xu2023ZeroshotOC, kang2024vlcounter}. These recent models
often do not match the performance of exemplar specified models, but
the open-vocabulary and language give new capabilities, such as only counting
objects in specific regions of the image~\cite{REC24}.

\section{\dataset Dataset}
\label{sec:dataset}

In this section, we introduce \dataset, a large scale video repetition counting dataset with text descriptions.
We first overview the statistics and content of the dataset, providing qualitative examples to illustrate its
diversity. Then in Section~\ref{sec:curation}, we describe how the dataset was built.

\begin{table}
\setlength{\tabcolsep}{0.3em}
\centering
\footnotesize{
    \begin{tabular}{@{}c|c|r|r|rcc|rcr|rrrr}
    \textbf{\dataset } & {\textbf{Dataset}}  & {\textbf{\# of}} & {\textbf{\# of}} &  \multicolumn{3}{c|}{\textbf{Duration (s)}} & \multicolumn{3}{c|}{\textbf{Count}} & \multicolumn{4}{c}{\textbf{Text}} \\
    \cline{3-14}
     \textbf{Source} & \textbf{split}  & {\textbf{videos}} & {\textbf{annos}} & \textbf{Avg} & \textbf{Min} & \textbf{Max} & \textbf{Avg} & \textbf{Min} & \textbf{Max} & \textbf{Avg} & \textbf{Min} & \textbf{Max}  & \textbf{Voc size}\\
    \midrule
     \multirow{ 3}{*}{\textbf{Kinetics} } & \textbf{ train} & 18127 & 19520 & 3.4 & 0.04 & 10.0 & 4.5 & 2 & 62 & 8.0 & 2 & 19 & 2468 \\
    & \textbf{consistent train} & 1111 & 2230 & 3.5 & 0.2 & 10.0 & 4.9 & 2 & 50 & 6.3 & 2 & 15 & 740 \\
     & \textbf{test} & 3760 & 7563 & 3.5 & 0.1 & 10.0 & 5.0 & 2 & 46 & 6.1 & 2 & 15 & 1163\\
    
    \midrule
    
    \multirow{ 3}{*}{\textbf{Ego4D} } & \textbf{train} & 41973 & 96336 & 2.5 & 0.03 & 10.0 & 4.7 & 2 & 121 & 7.2 & 1 & 23 & 3314 \\
    & \textbf{consistent train} & 33370 & 70179 & 2.5 & 0.1 & 10.0 & 4.7 & 2 & 71 & 7.2 & 1 & 21 & 2866\\
    & \textbf{test} & 8692 & 20409 & 2.5 & 0.10 & 10.00 & 4.8 & 2 & 61 & 7.2 & 1 & 21 & 1826 \\

    \end{tabular}
    \vspace{0.15em}
    }
\caption{\textbf{Detailed statistics of \dataset-Ego4D and \dataset-Kinetics datasets}} 
\label{tab:dataset_stats_splits}
\end{table}

\subsection{Dataset overview}

Our dataset is composed of two subsets: \dataset-Ego4D~\cite{grauman2022ego4d} and \dataset-Kinetics~\cite{kay2017kinetics,carreira2019short}, capturing both first-person and third-person views of human activities, respectively.

\begin{figure*}[t]
\footnotesize
\begin{center}
   \includegraphics[width=0.9\linewidth]{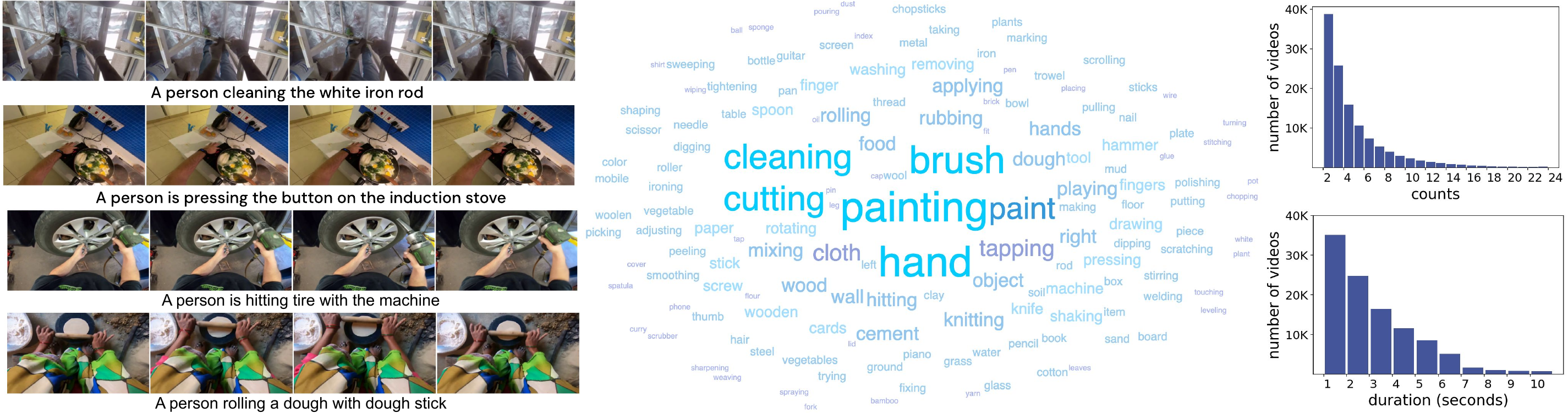}\\
  {\small (a) Example repetition video clips \qquad  (b) Word-clouds from repetition descriptions \hfill(c) Repetition statistics}
\end{center}
   \caption{\textbf{\dataset-Ego4D.} We display some example clips with free-form text descriptions in (a), the word-cloud visualisation of descriptions in (b), and repetition count and duration statistics in (c).}
\label{fig:ego4d_stats}
\end{figure*}

{\bf \dataset-Ego4D}. This subset is generated from publicly available Ego4D~\cite{grauman2022ego4d} videos. These videos are comprised of egocentric videos recorded by humans 
in a wide range of geographical locations and contextual settings. The details of the curation process are explained in Section~\ref{sec:curation} which results in a dataset of $50,665$ videos with free-form text descriptions.  Table~\ref{tab:dataset_stats_splits} displays the detailed train/test split statistics. Figure~\ref{fig:ego4d_stats}(a) demonstrates example repetitive video clips and their text annotations. Figure~\ref{fig:ego4d_stats}(c) presents the detailed statistics of repetition counts and durations. Note that there are a significant number of temporal repetitions with two cycles, which brings an additional challenge as recognising a repetitive pattern with only two cycles could be much harder compared to more cycles. Figure~\ref{fig:ego4d_stats}(b) visualises the word-cloud generated from the text descriptions which displays the diversity of human behaviours captured in our dataset. ``Person", ``man", ``woman", ``moving" words and stop-words are removed from the word-cloud to flush out details in the dataset.

\begin{figure*}[t]
\footnotesize
\begin{center}
   \includegraphics[width=0.9\linewidth]{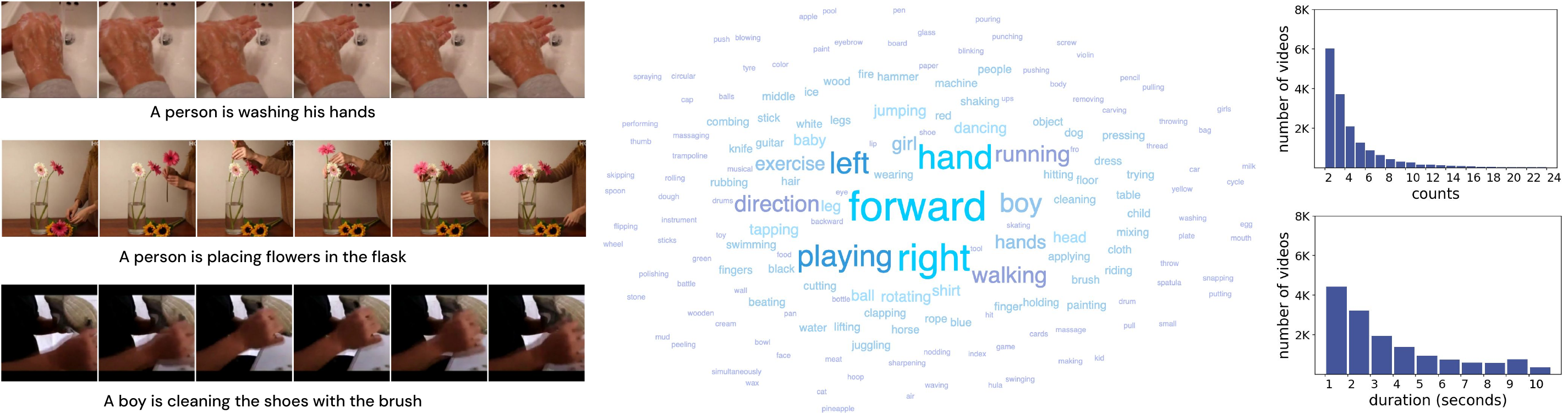}\\
   {\small (a) Example repetition video clips \qquad (b) Word-clouds from repetition descriptions \hfill(c) Repetition statistics}
\end{center}
   \caption{\textbf{\dataset-Kinetics.} We display some example clips with free-form text descriptions in (a), the word-cloud visualisation of descriptions in (b), and repetition count and duration statistics in (c). }
\label{fig:kinetics_stats}
\end{figure*}

{\bf \dataset-Kinetics}. This subset is generated from the publicly available Kinetics~\cite{smaira2020short} dataset similar to the Countix~\cite{dwibedi2020counting} dataset. However, instead of using a small number of categories with higher chance of repetitive behaviour (i.e.\ Countix), we annotate samples from the entire Kinetics dataset, hence the diversity is even more increased in our dataset.
This subset has $21,887$ videos with free-form text descriptions. The train/test statistics are displayed in Table~\ref{tab:dataset_stats_splits}, and visual examples, word-clouds and statistics of repetition counts and durations are demonstrated in Figure~\ref{fig:kinetics_stats} (a), (b), and (c), respectively. 

Comparing word-clouds in the two subsets, we can recognise different word distributions where \dataset-Ego4D focuses more on close-up human hand behaviours (e.g.\ painting, cutting, cleaning), tools (e.g.\ brush, hammer) and objects (e.g.\ food, wood, dough) whereas~\dataset-Kinetics has more emphasis on full-body human activities viewed with certain distance (e.g.\ walking, exercise, swimming, running) and directions (e.g.\ forward, right, left). 

\begin{figure*}[t]
\begin{center}
   \includegraphics[width=1.0\linewidth]{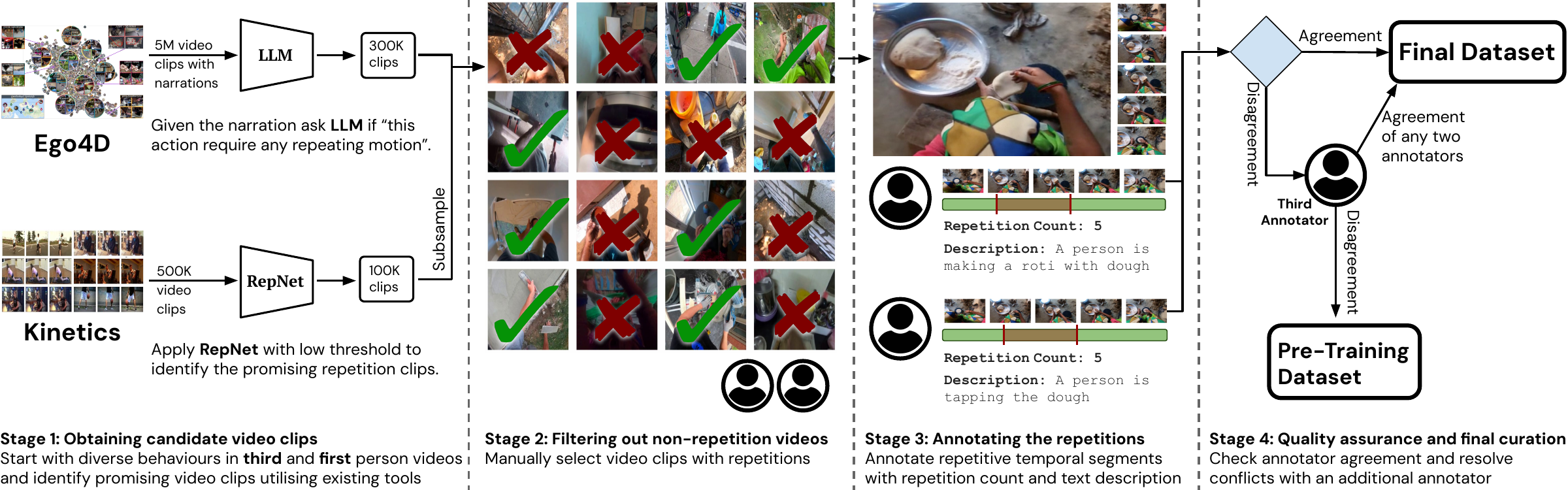}
\end{center}
   \caption{\textbf{Dataset Curation}. We explain how we construct \dataset dataset in four stages.}
\label{fig:dataset_curation}
\end{figure*}

\subsection{Dataset Curation}
\label{sec:curation}

In this section we describe the curation process: how
candidate video clips were obtained from Ego4D and Kinetics, and then the
processing pipeline that was used to select and annotate the candidates
and clean up the dataset. The overall curation process is illustrated in Figure~\ref{fig:dataset_curation}.

\begin{table}
    \footnotesize{
    \begin{subtable}{0.5\textwidth}
    \centering
    \begin{tabular}{l|r}
    \textbf{Ego4D Processing Stage } & \textbf{\# of clips}\\
    \midrule
        All  & 5,025,980 \\
        Chosen by LLM~\cite{chung2022scaling} & 889,424\\
        Remove overlapping clips & 308,514\\
        Sent for Manual Labeling & 100,000\\
        Valid Repetition Present & 52,813\\
        Consistent Count and Segment & 43,599 \\
    \end{tabular}
    \caption{Ego4D
    }
    ~~~
    \label{tab:ego4d_stages}
    \end{subtable}
     \begin{subtable}{0.5\textwidth}
    \centering
    \begin{tabular}{l|r}
    \textbf{Kinetics Processing Stage } & \textbf{\# of clips}\\
    \midrule
        All  & 650,317 \\
        Chosen by RepNet~\cite{dwibedi2020counting} & 351,751\\
        Sent for Manual Labeling & 54,665\\
        Valid Repetition Present & 21,887\\
        Consistent Count and Segment & 4,871 \\
    \end{tabular}
    \caption{Kinetics
    }
    \label{tab:kinetics_stages}
    \end{subtable}
    }
    \caption{Number of Clips at various stages of data curation.}
\end{table}

\noindent\textbf{Stage 1: Obtaining candidate video clips}

As repetitions are pervasive in daily
activities, we obtain clips with repetitions from existing video
datasets -- there is no need to collect a new
dataset of videos that just have repetitions. In particular, we source clips from two datasets: 
Ego4D~\cite{grauman2022ego4d} and Kinetics700v2~\cite{carreira2019short}. Ego4D consists of daily activity videos
collected from an egocentric viewpoint, while the Kinetics dataset consists of diverse human activities
collected from the web.  Both datasets are known for their diversity -- Ego4D, due to multiple contributors and geographical variation; Kinetics, because every instance for a class is sourced from a different YouTube video.
In terms of video length, we focus only on
the clips that are 10s long. While Kinetics videos are 10s by
design, for Ego4D, we sample 10s centered around the narration
timestamps to generate 10s clips. Two different methods are used to choose candidate clips with repetitions. Our goal at this stage is high diversity and high recall; we are less concerned with precision at this stage as subsequently we will manually filter out clips that do not have repetitions.

\noindent\textit{\textbf{RepNet based candidates.}} To select candidate clips from Kinetics we use the RepNet model
introduced in~\cite{dwibedi2020counting}. RepNet is a class agnostic
counting model and so can be applied to other classes/videos in
Kinetics. There is a concern that using a model to select candidate
videos will reduce diversity as only repetitions `seen through the
eyes' of RepNet will be included. There are two reasons why this is not such a concern as might first be thought:
first, RepNet has a `temporal self-similarity matrix' as a bottleneck. This means that if low level features are showing any repetition, then they will be picked out in the self-similarity matrix, irrespective of whether RepNet has seen instances of this repetition in training; Second, we use a low threshold ($0.25$) on the repetition classifier score to encourage diversity. We then apply manual filtering to confirm valid repetitions. We find that this filtering step results in $\sim 54$ candidate repeating videos for every 100 clips in the Kinetics dataset (see Table~\ref{tab:ego4d_stages}). While RepNet is a reliable repetition detector for non-egocentric videos (like the ones present in Kinetics), it is not as effective on egocentric videos of the Ego4D dataset. It often counts head and camera movements as repetitions.

\noindent\textit{\textbf{LLM based candidates.}}
For the Ego4D videos, we use a language model to select candidate
videos which may have repetitions. Ego4D provides a human supplied narration at a given timestamp. We choose a temporal window of 10 seconds around the narration timestamp as a candidate clip. We use only the narrations as
input for the LLM. The task of the LLM is to predict from the
narration alone whether there might be a repeating motion or not.
We use FlanT5~\cite{chung2022scaling} model for this step. We also ignore any walking
motion to focus more on tasks being performed by the person with their
hands as a disproportionately large number of clips of Ego4D involve
walking. The prompt provided to the LLM is described as follows: \textit{"Narration: <EGO4D Narration>. Q: Does the action described in above narration require any repeating motion? If the action is walking say no. A:"} \noindent Some example narrations classified as having repetitions are: \textit{\begin{enumerate*}
    \item C cuts the tree with the chainsaw in his right hand.
    \item The woman Y operates a phone with her hands.
    \item C peels the potato with the knife.
    \item The man Y climbs down the stairs.
    \item C mixes the paint on the paint box with the brush.
\end{enumerate*}} Some example narrations classified as not having repetitions are: \textit{\begin{enumerate*}
\item The man Y drops the boots on the floor.
\item C looks around the room.
\item C opens the garage door.
\item The man Z passes a bag to his left hand. 
\item C enters a room.
\end{enumerate*}}
We find that this steps results in $\sim 18$ candidate repeating videos for every 100 clips in the Ego4D dataset (see Table~\ref{tab:ego4d_stages}). 

\noindent\textbf{Stage 2: Filtering valid candidates}

The objective of this stage is to determine if the candidate clips actually contain repetitions or not according to human annotators. We ask them not to label very common repetitions like walking, breathing, etc. Beyond that the annotators were free to choose which repetition to focus on so that we can get a diverse set of actions. We prepared a web-based annotation tool that displays one video clip at a time to an annotator, and asked them if the clip had a repeating action in it or not. To further increase robustness we show each video to two different annotators and check the agreement between them. This stage is important as both the LLM and RepNet have their own sets of false positives. Since the LLM based filtering does not use the video and only the text, a common pitfall is that while the action in the narration generally involves repetitions at some point (\textit{using a phone}), such repetitions were not there or are not clear in the video. Common false positives from RepNet include clips with walking or repetitive camera movements.

\noindent\textbf{Stage 3: Annotating the repetitions}

The objective of this stage is to annotate the repetition in detail in the valid candidates.
Our web-based annotation tool enables users to smoothly go back and forth in the video and enter answers to four questions:
\begin{enumerate*}[(a)]
    \item What is the repeating action?
    \item When does the repetition start?
    \item When does the repetition end?
    \item How many times does the repetition happen?
\end{enumerate*}

Through these questions we obtain the temporal interval of the contiguous repetitions, the repetition count, and a free-form text description of the repetition. For this stage also, initially we use two annotators. If multiple repetitions happen in a video we ask the annotators to annotate only the earliest repetition in the clip. This is done to ensure consistent labeling.

\noindent\textbf{Stage 4: Quality assurance and final curation}

Finally in the last stage, we resolve disagreements through a third annotator. For any video if the existing two annotators agreed on the temporal repetition segment ($\ge  50\%$ IOU between the two segments) but disagreed in the repetition count by more than 1, we re-annotate the video again with a new annotator. If there is agreement between any two of the three annotators we include the sample in the final dataset. We retain all annotations. We call the videos with at least 2 agreeing annotators as \textit{consistent}. All videos in the test split are consistent while a subset of the videos in the train split are consistent. We do not get rid of the non-consistent in case training on more data with noisy labels is helpful in the future. We then create train/test splits from these videos. We use a split of 80\% train and 20\% test. The test split of \dataset-Kinetics is from the test split of Kinetics-700. We hope the large number of test videos is useful in future evaluations.

\section{\modelname: A Conditional Counting Model}
\label{sec:model}

\begin{figure*}[t]
\centering
   \includegraphics[width=1.0\linewidth]{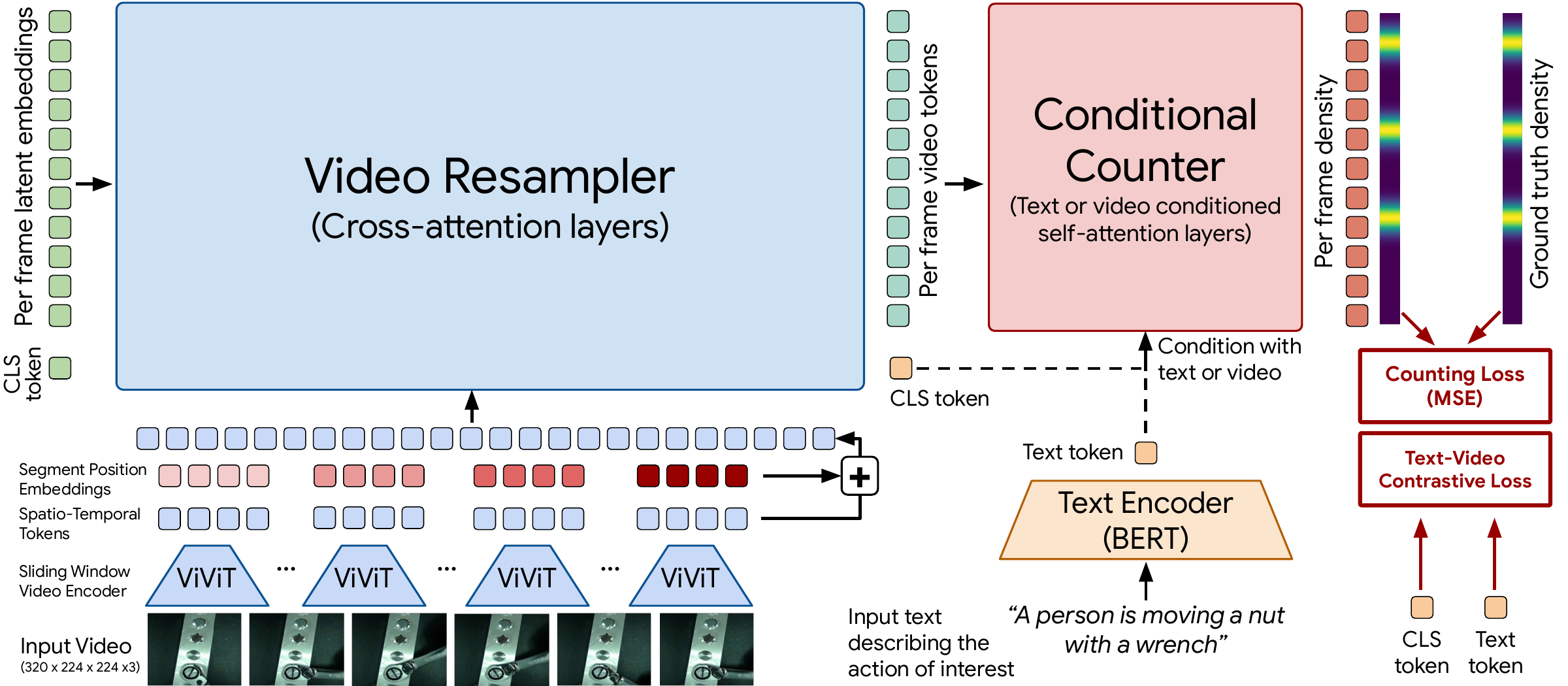}
   \caption{\textbf{\modelname Architecture.} An input video goes through a spatio-temporal video encoder (bottom left) and then resampled through the Video Resampler module to generate per frame video tokens. These per frame video tokens run through the Conditional Counter to generate the per-frame densities which are used to compute the final count. The Conditional Counter is either conditioned with CLS token enabling class-agnostic counting or text token enabling text-conditioned counting.}
\label{fig:architecture}
\end{figure*}

Our objective is to train a baseline model to show the effectiveness of the collected data in training counting models. We train a single model that can perform temporal repetition counting with or without text conditioning. Below we will discuss all the components of our model shown in Figure~\ref{fig:architecture}. 

\noindent\textbf{Input.} The model takes a video $V = [f_1, f_2, ... f_N]$ as input where $N$ is the number of frames and $f_i$ denotes a single frame. In this work, consider processing $N=320$ frames at once. This exceeds the maximum length of a video in the \dataset which is of 300 frames.

\noindent\textbf{Sliding Window Video Encoder.} First, we pass non-overlapping segments of length $K$ frame samples at a stride of $S$ of the video through a video segment encoder.  This produces $\frac{N}{K \times S}$ segments. Say each segment passed through the video segment encoder produces $n$ tokens. Since, we want to keep all the spatio-temporal tokens for processing, we end up with  $\frac{N n}{K \times S}$ tokens. Specifically we use a Kinetics-400 pre-trained ViViT~\cite{arnab2021vivit} model that can take $K=16$ frames at a stride of $S=4$ frames as input pre-trained with Masked Auto-encoder~\cite{tong2022videomae} loss. For videos of length $N=320$ where each frame is of size $224\times224\times3$, we end up with $\frac{320}{16 \times 4}=5$ segments and $5 \times 1568 = 7840$ tokens. Further processing using all $7840$ spatio-temporal tokens with self-attention transformer layers would result in extremely large attention matrices. Hence we use a video resampler module (described below) using cross-attention layers which are more memory efficient.

\noindent{\textbf{Video Resampler.}} We add a segment position encoding to the spatio-temporal tokens to let the transformer know the temporal ordering of the segments. We then pass the spatio-temporal tokens with the position embedding to 2 cross-attention transformer decoder layers. The goal of these layers is to perform two objectives: i) perform temporal reasoning ii) reduce the number of tokens required for further processing. We learn $N$ latent vectors multiplied with positional embeddings as queries for the cross-attention layers. Each token corresponds to a single frame in the video. This correspondence is established through the loss function that is applied at a per-frame level.

\noindent{\textbf{Conditional Counter.}} We use an Adaptive LayerNorm (AdaLN)~\cite{peebles2023scalable} self-attention transformer layer as our counting module. This decoder composed of 2 layers takes per-frame video tokens as input queries and a conditioning token used to modulate the LayerNorm outputs used in the transformer layers. The conditioning token can either be the the CLS token output of the Video Resampler in which case the model performs class-agnostic counting, or alternatively we can also use text to condition the counter to only count actions if they are relevant to the text used in the description. We use text tokens from a frozen BERT~\cite{bert} model as our textual conditioning. Conditional Counter is expected to produce a per-frame density with peaks at the temporal mid-points of each period of the relevant repeating action and zeros elsewhere. Summing up the per-frame densities give us the count for the whole video.

\noindent{\textbf{Losses.}} We mainly have two loss functions: (a) a \emph{counting loss} that minimizes the difference between predicted and expected per frame densities (i.e.\ mean squared error (MSE) loss), and (b) a \emph{text-video contrastive loss} which trains the CLS token to be used instead of text when text is not available. For every training sample, we minimize the counting loss in three different scenarios: (i) use only video, where CLS token is used in Conditional Counter; (ii) use text and video, where text token is used in Conditional Counter, (iii) use video with a mismatched text, where text token is used in Conditional Counter and the Ground Truth density is set to a zero vector. We also train a text-video contrastive loss (CLIP style) in order to align the CLS token output (from the Video Resampler) with the text token (from the Text Encoder). This provides a semantic supervision to the Video Resampler and enables the CLS token to be used instead of text token (when not available) in Conditional Counter.
\section{Experiments}
\label{sec:experiments}

We consider two tasks: class-agnostic counting, and text-conditioned counting. Class-agnostic counting refers to the counting of any repetitions in a video, whereas in text-conditioned only repetitions with the class specified by the text are counted, and other repetitions are ignored. In addition to the counting task, we also assess the task of repetition segmentation, which is to determine when the contiguous repetitions start and end. 

\noindent\textbf{Evaluation Metrics.} We use the following metrics in order to evaluate different tasks related to repetition counting: \begin{enumerate*}
\item \textit{Counting.} As is common in repetition counting literature we use off-by-one error (OBOE) and mean absolute error (MAE) to measure if the model can accurately count repetitions or not. We also report off-by-zero error (OBZE) and root mean squared error (RMSE) which are newer repetition counting metrics introduced in~\cite{sinha2024shot}. \item \textit{Repetition Segmentation.} In order to measure if models are aware of when the repetition is happening in a video we measure the intersection over union (IOU) between ground truth and predicted repeating segment. In case there are multiple repeating segments in one video we consider only the first/earliest segment to calculate this metric.
\end{enumerate*}

\begin{figure*}[t!]
\centering
   \includegraphics[width=\textwidth]{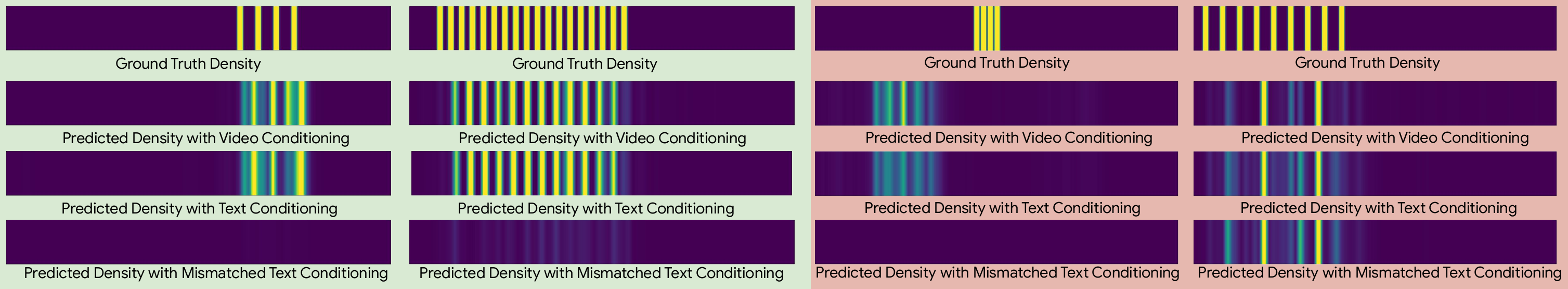}
   \caption{\textbf{Examples of correct and incorrect predictions by \modelname.}  More details in Sec~\ref{sec:results}}
\label{fig:predictions}
\end{figure*}

\begin{table}[t]
\setlength{\tabcolsep}{0.5em}
\footnotesize
{
\begin{tabular}{l|c|lllll|lllll}
   & & \multicolumn{5}{c|}{\dataset-Kinetics} & \multicolumn{5}{c}{\dataset-Ego4D}\\
\midrule
Model   &Cond. &\scriptsize MAE  {\tiny$\downarrow$ }& \scriptsize OBOE  {\tiny$\downarrow$ } & \scriptsize OBZE  {\tiny$\downarrow$ }& \scriptsize RMSE  {\tiny$\downarrow$ }& \scriptsize IOU {\tiny{\tiny$\uparrow$ }} & \scriptsize MAE  {\tiny$\downarrow$ } & \scriptsize OBOE  {\tiny$\downarrow$ }& \scriptsize OBZE  {\tiny$\downarrow$ }& \scriptsize RMSE  {\tiny$\downarrow$ }& \scriptsize IOU {\tiny$\uparrow$}\\
\midrule
RepNet~\cite{dwibedi2020counting}  & Video & 0.86  &  0.65    & 0.87      & 3.57  & 0.35 & 0.74     &   0.57    &   0.81    &     3.20   &  0.26 \\
\midrule
\modelname & Video & 0.39 &	0.41 &	0.75 &	2.18 &	0.45 &	0.35 &	0.34 &	0.71 &	1.60 &	0.45\\
\modelname & Text & 0.39 &	0.40 &	0.75 &	2.14 &	0.46 &	0.35 &	0.34 &	0.71 & 1.60 & 0.45 \\
\end{tabular}
}
\caption{Results on the \dataset dataset for Repetition Counting and Periodicity Localization tasks.}
\label{tab:cagctr}
\end{table}

\noindent\textbf{Results.}
\label{sec:results}
 We report the results in Table~\ref{tab:cagctr}. We find that a prior model RepNet on class-agnostic counting is unable to produce accurate counts or repetition segments on our dataset. This is mainly due to RepNet being trained on a cleaner data where videos mostly contain repeating actions. However, in our dataset only some parts of the video has repeating actions. Our baseline model \modelname that is trained on \dataset dataset performs much better in all metrics. There is still much room for improvement as the harder counting metric Off-by-Zero-Error is  0.75 for \dataset-Kinetics and 0.71 for \dataset-Ego4D. We also find a significant increase in the ability of the model to localize repeating segments in videos. For \dataset-Kinetics, average IOU increased from 0.3 to 0.45 and for \dataset-Ego4D average IOU increased from 0.26 to 0.45.

The second setting we study is the text-conditioned setup where the action of interest is provided as input to the model. We report the results in Table~\ref{tab:cagctr}. We find our model is able to count without any performance drop when conditioned with text. This shows the flexibility of the model in being used in two different settings. We also study what happens when a mismatched text is provided as conditioning input. On the \dataset-Kinetics dataset the off-by-zero error is 0.30 and on \dataset-Kinetics dataset is 0.29. This indicates that the model does pay attention to the text conditioning. However, more work is needed to improve the OBZE metric when there is a text mismatch.

We show examples of \modelname's predictions in Figure~\ref{fig:predictions}. On the left two columns (green), we show two examples where the model correctly counts and localizes the repetitions. Note how when we provide a mismatched text (bottom row) the density predictions are close to zero indicating the model is ignoring those mis-matched repetitions. On the right two columns (red) we provide examples of incorrect predictions. In the third column the model is counting repetitions in a different part of the video. In the fourth column, the model is not able to count properly. We also observe it is producing counts with a mismatched text conditioning.

\section{Limitations and Broader Impact}
\label{sec:limitations}

While we aim to scale the diversity of repetition counting datasets, we limit ourselves to clips that are 10s long as ensuring consistency of count and localization labels between annotators with such short clips was found to be challenging. Text-conditioned repetition counting in videos has broad implications for sports and health analytics, and vision based industrial process counting. However, there are concerns if the model is used for unethical surveillance or used without mitigating biases.

\section{Conclusion}
\label{sec:conclusion}

We introduce a large dataset of temporal repetitions in videos annotated with free-form text. This dataset contains videos captured from both egocentric and exocentric viewpoints.  We also train a baseline model \modelname that is able to count repetitions in videos in an end to end manner. We hope this dataset and the model will help the community in developing better temporal reasoning visual models.

{
    \small
    \bibliographystyle{abbrv}
    \bibliography{main}

\begin{thebibliography}{10}

\bibitem{AminiNaieni23}
N.~Amini-Naieni, K.~Amini-Naieni, T.~Han, and A.~Zisserman.
\newblock Open-world text-specified object counting.
\newblock In {\em BMVC}, 2023.

\bibitem{arnab2021vivit}
A.~Arnab, M.~Dehghani, G.~Heigold, C.~Sun, M.~Lu{\v{c}}i{\'c}, and C.~Schmid.
\newblock Vivit: A video vision transformer.
\newblock In {\em Proceedings of the IEEE/CVF international conference on
  computer vision}, pages 6836--6846, 2021.

\bibitem{azy2008segmentation}
O.~Azy and N.~Ahuja.
\newblock Segmentation of periodically moving objects.
\newblock In {\em 2008 19th International Conference on Pattern Recognition},
  pages 1--4. IEEE, 2008.

\bibitem{briassouli2007extraction}
A.~Briassouli and N.~Ahuja.
\newblock Extraction and analysis of multiple periodic motions in video
  sequences.
\newblock {\em IEEE transactions on pattern analysis and machine intelligence},
  29(7):1244--1261, 2007.

\bibitem{carreira2019short}
J.~Carreira, E.~Noland, C.~Hillier, and A.~Zisserman.
\newblock A short note on the kinetics-700 human action dataset.
\newblock {\em arXiv preprint arXiv:1907.06987}, 2019.

\bibitem{chung2022scaling}
H.~W. Chung, L.~Hou, S.~Longpre, B.~Zoph, Y.~Tay, W.~Fedus, Y.~Li, X.~Wang,
  M.~Dehghani, S.~Brahma, et~al.
\newblock Scaling instruction-finetuned language models.
\newblock {\em arXiv preprint arXiv:2210.11416}, 2022.

\bibitem{cutler2000robust}
R.~Cutler and L.~S. Davis.
\newblock Robust real-time periodic motion detection, analysis, and
  applications.
\newblock {\em IEEE Transactions on pattern analysis and machine intelligence},
  22(8):781--796, 2000.

\bibitem{REC24}
S.~Dai, J.~Liu, and N.-M. Cheung.
\newblock Referring expression counting.
\newblock In {\em CVPR}, 2024.

\bibitem{bert}
J.~Devlin, M.~Chang, K.~Lee, and K.~Toutanova.
\newblock {BERT:} pre-training of deep bidirectional transformers for language
  understanding.
\newblock {\em CoRR}, 2018.

\bibitem{low_shot}
N.~Djukic, A.~Lukezic, V.~Zavrtanik, and M.~Kristan.
\newblock A low-shot object counting network with iterative prototype
  adaptation.
\newblock {\em arXiv preprint arXiv:2211.08217}, 2022.

\bibitem{dwibedi2020counting}
D.~Dwibedi, Y.~Aytar, J.~Tompson, P.~Sermanet, and A.~Zisserman.
\newblock Counting out time: Class agnostic video repetition counting in the
  wild.
\newblock In {\em Proceedings of the IEEE/CVF conference on computer vision and
  pattern recognition}, pages 10387--10396, 2020.

\bibitem{Gong2022ClassAgnosticOC}
S.~Gong, S.~Zhang, J.~Yang, D.~Dai, and B.~Schiele.
\newblock Class-agnostic object counting robust to intraclass diversity.
\newblock In {\em ECCV}, 2022.

\bibitem{grauman2022ego4d}
K.~Grauman, A.~Westbury, E.~Byrne, Z.~Chavis, A.~Furnari, R.~Girdhar,
  J.~Hamburger, H.~Jiang, M.~Liu, X.~Liu, et~al.
\newblock Ego4d: Around the world in 3,000 hours of egocentric video.
\newblock In {\em Proceedings of the IEEE/CVF Conference on Computer Vision and
  Pattern Recognition}, pages 18995--19012, 2022.

\bibitem{hu2022transrac}
H.~Hu, S.~Dong, Y.~Zhao, D.~Lian, Z.~Li, and S.~Gao.
\newblock Transrac: Encoding multi-scale temporal correlation with transformers
  for repetitive action counting.
\newblock In {\em Proceedings of the IEEE/CVF Conference on Computer Vision and
  Pattern Recognition}, pages 19013--19022, 2022.

\bibitem{Jiang2023CLIPCountTT}
R.~Jiang, L.~Liu, and C.~Chen.
\newblock Clip-count: Towards text-guided zero-shot object counting.
\newblock In {\em Proceedings of the 31st ACM International Conference on
  Multimedia}, 2023.

\bibitem{kang2024vlcounter}
S.~Kang, W.~Moon, E.~Kim, and J.-P. Heo.
\newblock Vlcounter: Text-aware visual representation for zero-shot object
  counting.
\newblock In {\em Proceedings of the AAAI Conference on Artificial
  Intelligence}, volume~38, pages 2714--2722, 2024.

\bibitem{kay2017kinetics}
W.~Kay, J.~Carreira, K.~Simonyan, B.~Zhang, C.~Hillier, S.~Vijayanarasimhan,
  F.~Viola, T.~Green, T.~Back, P.~Natsev, et~al.
\newblock The kinetics human action video dataset.
\newblock {\em arXiv preprint arXiv:1705.06950}, 2017.

\bibitem{Lempitsky10b}
V.~Lempitsky and A.~Zisserman.
\newblock Learning to count objects in images.
\newblock 2010.

\bibitem{levy2015live}
O.~Levy and L.~Wolf.
\newblock Live repetition counting.
\newblock In {\em Proceedings of the IEEE international conference on computer
  vision}, pages 3020--3028, 2015.

\bibitem{li2023full}
J.~Li, B.~Chen, Z.~Wang, and H.~Liu.
\newblock Full resolution repetition counting.
\newblock {\em arXiv preprint arXiv:2305.13778}, 2023.

\bibitem{Lin_2022_BMVC}
W.~Lin, K.~Yang, X.~Ma, J.~Gao, L.~Liu, S.~Liu, J.~Hou, S.~Yi, and A.~Chan.
\newblock Scale-prior deformable convolution for exemplar-guided class-agnostic
  counting.
\newblock In {\em BMVC}, 2022.

\bibitem{Liu22}
C.~Liu, Y.~Zhong, A.~Zisserman, and W.~Xie.
\newblock Countr: Transformer-based generalised visual counting.
\newblock In {\em BMVC}, 2022.

\bibitem{Lu18}
E.~Lu, W.~Xie, and A.~Zisserman.
\newblock Class-agnostic counting.
\newblock In {\em ACCV}, 2018.

\bibitem{mundhenkLargeContextualDataset2016a}
T.~N. Mundhenk, G.~Konjevod, W.~A. Sakla, and K.~Boakye.
\newblock A large contextual dataset for classification, detection and counting
  of cars with deep learning.
\newblock In {\em ECCV}, 2016.

\bibitem{10.1007/978-3-031-20044-1_20}
T.~Nguyen, C.~Pham, K.~Nguyen, and M.~Hoai.
\newblock Few-shot object counting and detection.
\newblock In {\em ECCV}, 2022.

\bibitem{peebles2023scalable}
W.~Peebles and S.~Xie.
\newblock Scalable diffusion models with transformers.
\newblock In {\em Proceedings of the IEEE/CVF International Conference on
  Computer Vision}, pages 4195--4205, 2023.

\bibitem{dave}
J.~Pelhan, A.~Lukežič, V.~Zavrtanik, and M.~Kristan.
\newblock Dave – a detect-and-verify paradigm for low-shot counting.
\newblock 2024.

\bibitem{pogalin2008visual}
E.~Pogalin, A.~W. Smeulders, and A.~H. Thean.
\newblock Visual quasi-periodicity.
\newblock In {\em 2008 IEEE Conference on Computer Vision and Pattern
  Recognition}, pages 1--8. IEEE, 2008.

\bibitem{m_Ranjan-etal-CVPR21}
V.~Ranjan, U.~Sharma, T.~Nguyen, and M.~Hoai.
\newblock Learning to count everything.
\newblock In {\em CVPR}, 2021.

\bibitem{runia2018real}
T.~F. Runia, C.~G. Snoek, and A.~W. Smeulders.
\newblock Real-world repetition estimation by div, grad and curl.
\newblock In {\em Proceedings of the IEEE conference on computer vision and
  pattern recognition}, pages 9009--9017, 2018.

\bibitem{Shi2022RepresentCA}
M.~Shi, H.~Lu, C.~Feng, C.~Liu, and Z.~CAO.
\newblock Represent, compare, and learn: A similarity-aware framework for
  class-agnostic counting.
\newblock In {\em CVPR}, 2022.

\bibitem{sinha2024shot}
S.~Sinha, A.~Stergiou, and D.~Damen.
\newblock Every shot counts: Using exemplars for repetition counting in videos,
  2024.

\bibitem{smaira2020short}
L.~Smaira, J.~Carreira, E.~Noland, E.~Clancy, A.~Wu, and A.~Zisserman.
\newblock A short note on the kinetics-700-2020 human action dataset.
\newblock {\em arXiv preprint arXiv:2010.10864}, 2020.

\bibitem{soomro2012ucf101}
K.~Soomro, A.~R. Zamir, and M.~Shah.
\newblock Ucf101: A dataset of 101 human actions classes from videos in the
  wild.
\newblock {\em arXiv preprint arXiv:1212.0402}, 2012.

\bibitem{tong2022videomae}
Z.~Tong, Y.~Song, J.~Wang, and L.~Wang.
\newblock Videomae: Masked autoencoders are data-efficient learners for
  self-supervised video pre-training.
\newblock {\em Advances in neural information processing systems},
  35:10078--10093, 2022.

\bibitem{tsai1994cyclic}
P.-S. Tsai, M.~Shah, K.~Keiter, and T.~Kasparis.
\newblock Cyclic motion detection for motion based recognition.
\newblock {\em Pattern recognition}, 27(12):1591--1603, 1994.

\bibitem{vlachos2005periodicity}
M.~Vlachos, P.~Yu, and V.~Castelli.
\newblock On periodicity detection and structural periodic similarity.
\newblock In {\em Proceedings of the 2005 SIAM international conference on data
  mining}, pages 449--460. SIAM, 2005.

\bibitem{Xu2023ZeroshotOC}
J.~Xu, H.~Le, V.~Nguyen, V.~Ranjan, and D.~Samaras.
\newblock Zero-shot object counting.
\newblock In {\em CVPR}, 2023.

\bibitem{yangClassagnosticFewshotObject2021}
S.-D. Yang, H.-T. Su, W.~H. Hsu, and W.-C. Chen.
\newblock Class-agnostic few-shot object counting.
\newblock 2021.

\bibitem{yao2023poserac}
Z.~Yao, X.~Cheng, and Y.~Zou.
\newblock Poserac: Pose saliency transformer for repetitive action counting.
\newblock {\em arXiv preprint arXiv:2303.08450}, 2023.

\bibitem{You_2023_WACV}
Z.~You, K.~Yang, W.~Luo, X.~Lu, L.~Cui, and X.~Le.
\newblock Few-shot object counting with similarity-aware feature enhancement.
\newblock 2023.

\bibitem{zhang2020context}
H.~Zhang, X.~Xu, G.~Han, and S.~He.
\newblock Context-aware and scale-insensitive temporal repetition counting.
\newblock In {\em Proceedings of the IEEE/CVF Conference on Computer Vision and
  Pattern Recognition}, pages 670--678, 2020.

\bibitem{zhang2021repetitive}
Y.~Zhang, L.~Shao, and C.~G. Snoek.
\newblock Repetitive activity counting by sight and sound.
\newblock In {\em Proceedings of the IEEE/CVF Conference on Computer Vision and
  Pattern Recognition}, pages 14070--14079, 2021.

\end{thebibliography}
}

\appendix
\appendix

\section{Dataset Release Details}

The dataset is being released as 2 JSON files available to download at these links hosted on Google Cloud: \href{https://storage.googleapis.com/semantic_repetitions/ovr_ego4d_release.json}{OVR-Ego4D} and \href{https://storage.googleapis.com/semantic_repetitions/ovr_kinetics_release.json}{OVR-Kinetics}. Both datasets have the following structure:

\begin{enumerate}
\item \verb|description| - Description of the activity that is being repeated.
\item \verb|start_time| - Time in seconds when repetition starts in the video. (Time is calculated as offset from the chosen clip)
\item \verb|end_time| - Time in seconds when repetition ends in the video.  (Time is calculated as offset from the chosen clip)
\item \verb|count| - The number of times the action was repeated in the video.
\item \verb|split| - Part of Train or Test split
\item \verb|inter_rater_agreement| - Annotation where the labels have 50\% IOU and Off-by-One count of 1 with the other annotations. Used to ensure test set has annotations where at least 2 raters agree.
\end{enumerate}

For Kinetics, we provide the \verb|video_id| to identify the unique clip sampled at 25fps. . For Ego4D, we provide a combination of \verb|video_id| and \verb|narration_timestamp_secs| to mark the unique clip as multiple repeating clips exist in a single Ego4D video. Each clip in Ego4D is sampled at \verb|narration_timestamp_secs| $\pm~5$s at 30fps.

\begin{figure*}[t]
\begin{center}
  \includegraphics[width=1.0\linewidth]{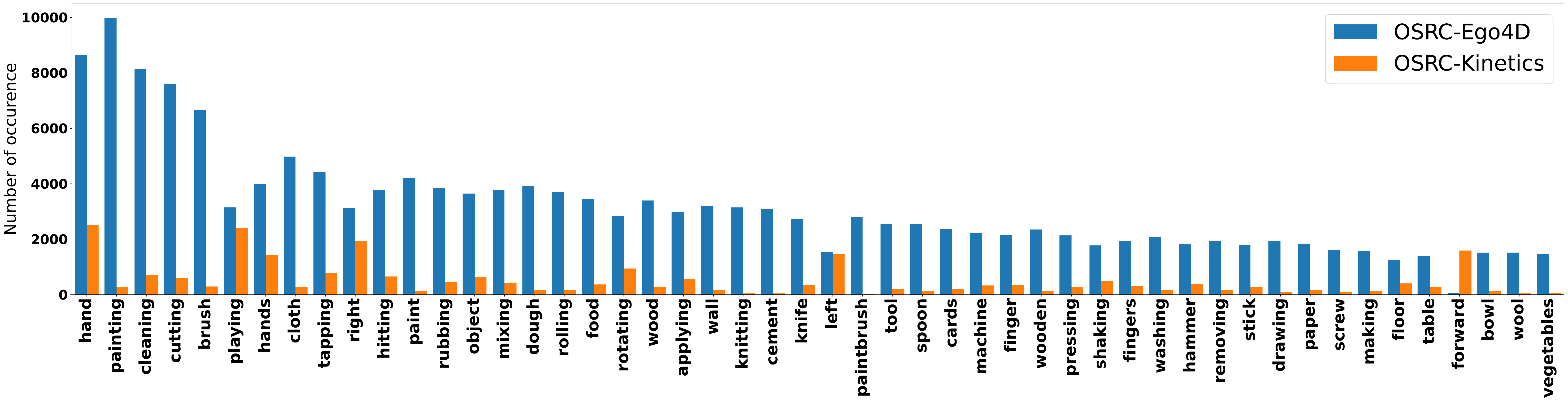}
\end{center}
  \caption{\textbf{Frequency distribution of words common to both \dataset-Ego4D and \dataset-Kinetics}.  Better viewed with zoom.}
\label{fig:dataset_word_dist1}
\end{figure*}

\begin{figure*}[htp]
\begin{center}
  \includegraphics[width=1.0\linewidth]{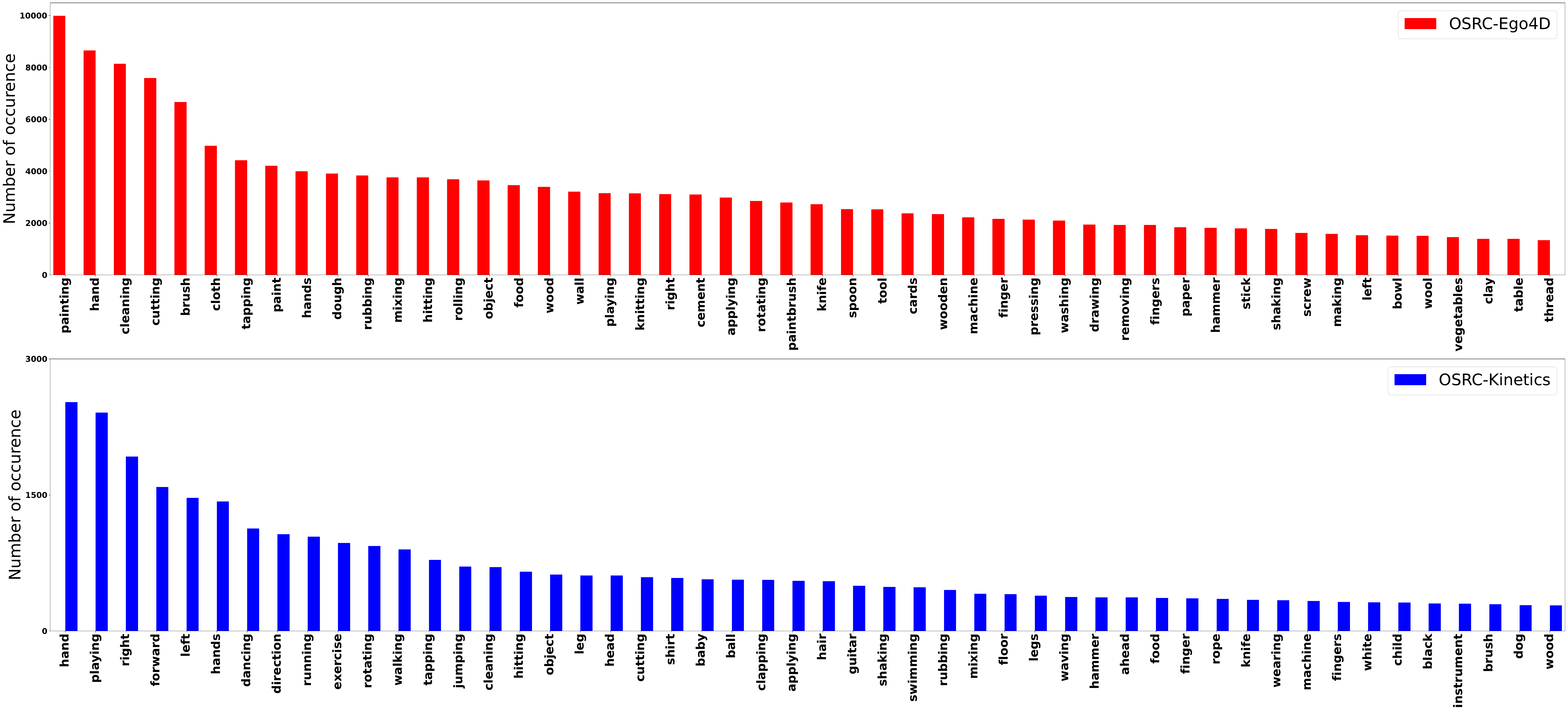}
\end{center}
  \caption{\textbf{Frequency distribution of words in \dataset-Ego4D and \dataset-Kinetics}. Better viewed with zoom. 
}
\label{fig:dataset_word_dist2}
\end{figure*}

\section{Word Distribution}
In Figure~\ref{fig:dataset_word_dist1}, we plot the distribution of the most common 50 words between the two datasets ordered by the sum of occurrence in both datasets. Note the difference in distribution in chosen verbs and nouns in first-person (\dataset-Ego4D) and third-person (\dataset-Kinetics) videos. Stop-words and \textit{``person",``man",``woman",``boy",``girl',``moving"} words are removed to concentrate on the differences between the datasets.  In Figure~\ref{fig:dataset_word_dist2}, we plot the word frequency distribution individually over the two datasets to show how the common verbs and nouns differ in both datasets.

\section{\modelname Details}
\noindent\textbf{Architecture.} We used two Transformer Cross-Attention layers as our video resampler and two AdaLN~\cite{peebles2023scalable} Self-Attention layers as our conditional counter. Both of these modules have a  12 attention heads with 768 dims for the self-attention layers and  3072 dims for the MLP layers.

\noindent\textbf{Training.} We train \modelname with a batch size of $64$ videos of $320$ frames with learning rate of $10^{-4}$ with AdamW optimizer for $50K$ steps. The model is trained to predict per-frame densities of the mid-point of repeating periods~\cite{sinha2024shot} which can be summed to provide the count. The densities are scaled by 100 before applying the MSE loss. The full loss function is defined as follows:

\begin{align*}
    L = \frac{L^{class-agnostic}_{count}  +  L^{class-aware}_{count} +  L^{mismatch-class-aware}_{count} + 10 * L_{contrastive}}{4}
\end{align*}

\end{document}